\documentclass[letterpaper, 10 pt, journal]{packages/IEEEtran}

\IEEEoverridecommandlockouts                              

\usepackage{times}
\usepackage{epsfig}
\usepackage{graphicx}
\usepackage{amsmath}
\usepackage{amssymb}
\usepackage{subfigure} 
\usepackage[ruled]{algorithm}
\usepackage{stackengine}
\usepackage{scalerel}
\usepackage{textcomp}
\usepackage{stfloats}
\usepackage{lettrine}
\usepackage{url}
\usepackage{soul}
\usepackage[ruled]{algorithm}
\usepackage{algpseudocode}
\usepackage{ctable}
\usepackage{cite}
\usepackage[export]{adjustbox}
\usepackage{import}
\usepackage{pgf}
\usepackage[bookmarks=true]{hyperref}

\UseRawInputEncoding

\makeatletter
\let\NAT@parse\undefined
\makeatother
\usepackage[numbers]{natbib}

\usepackage{packages/sparo_acronyms}
\usepackage{packages/sparo_math}
\usepackage{packages/sparo_SIunits}
\usepackage{packages/sparo_misc}
\usepackage{multirow}

\usepackage{soul,color}
\usepackage{verbatim} 

\usepackage{xcolor}
\newcommand{\bl}[1]{{\textcolor{black}{#1}}}

\definecolor{gl}{HTML}{008000}

\DeclareMathOperator*{\argmin}{argmin}

\usepackage[symbol]{footmisc}

\title{ReFeree: Radar-Based Lightweight and Robust Localization using Feature and Free space
}

\author{Hogyun Kim$^{1*}$ Byunghee Choi$^{1*}$, Euncheol Choi$^{1}$, and Younggun Cho$^{1\dagger}$
\thanks{Manuscript received: July, 24, 2024; accepted: September, 18, 2024. This letter was recommended for publication by Associate Editor J. Civera upon evaluation of the reviewers' comments. This work was supported by the National Research Foundation of Korea(NRF) grant (No. RS-2023-00302589 and No.2022R1A4A3029480) and Institute of Information \& communications Technology Planning \& Evaluation grant (No.2022-0-00448) funded by the Korea government(MSIT). } 
\thanks{$^{1*}$Hogyun Kim, $^{1*}$Byunghee Choi, $^{1}$Euncheol Choi, and $^{1\dagger}$Younggun Cho are with the Electrical and Computer Engineering, Inha University, Incheon, South Korea {\tt\small [hg.kim, bhbhchoi, 22231346]@inha.edu, yg.cho@inha.ac.kr} \break \hfill
        (*) represents equal contribution. }%
\thanks{Digital Object Identifier (DOI): see top of this page.}
}

\begin{document}

\maketitle

\begin{abstract}
Place recognition plays an important role in achieving robust long-term autonomy.
Real-world robots face a wide range of weather conditions (e.g. overcast, heavy rain, and snowing) and most sensors (i.e. camera, LiDAR) essentially functioning within or near-visible electromagnetic waves are sensitive to adverse weather conditions, making reliable localization difficult.
In contrast, radar is gaining traction due to long electromagnetic waves, which are less affected by environmental changes and weather independence.
In this work, we propose a radar-based lightweight and robust place recognition.
We achieve rotational invariance and lightweight by selecting a one-dimensional ring-shaped description and robustness by mitigating the impact of false detection utilizing opposite noise characteristics between free space and feature.
In addition, the initial heading can be estimated, which can assist in building a SLAM pipeline that combines odometry and registration, which takes into account onboard computing.
The proposed method was tested for rigorous validation across various scenarios (i.e. single session, multi-session, and different weather conditions).
In particular, we validate our descriptor achieving reliable place recognition performance through the results of extreme environments that lacked structural information such as an OORD dataset.
Our supplementary materials and code are available at \texttt{\url{https://sites.google.com/view/referee-radar}}.

\end{abstract}

 
\section{Introduction}
\IEEEPARstart{W}{ithout} prior knowledge, place recognition allows the robot's pose to be estimated on a map from onboard sensor measurements.
Especially, this task is crucial for robot navigation, multi-robot mapping, and \ac{SLAM}.
Unlike vision-based place recognition \cite{kim2023robust}, and \ac{LiDAR}-based place recognition \cite{he2016m2dp, kim2018scan, uy2018pointnetvlad, kim2024narrowing} affected by adverse weather, range detection and ranging (radar)-based place recognition has drawn attention to its robustness~\cite{jang2023raplace, gadd2024open}.

However, these methods focus on compact scene compression and rotational or translational invariance. 
Thus, they do not incorporate semi-metric information into the description, making it difficult to correct the accumulated pose error through initial heading estimation for the reverse loop.
In addition, the time it takes to generate their descriptors or compare the similarity between the two descriptors can hinder place recognition on an onboard computer.
There are some attempts to apply place recognition methods for 3D point clouds to 2D point clouds from radar, but reliability degradation is inevitable as dimensionality is reduced \cite{hong2022radarslam}.

\begin{figure}[t]
	\centering
	\def\width{0.49\textwidth}%
    {%
        \includegraphics[clip, trim= 0 80 0 70, width=\width]{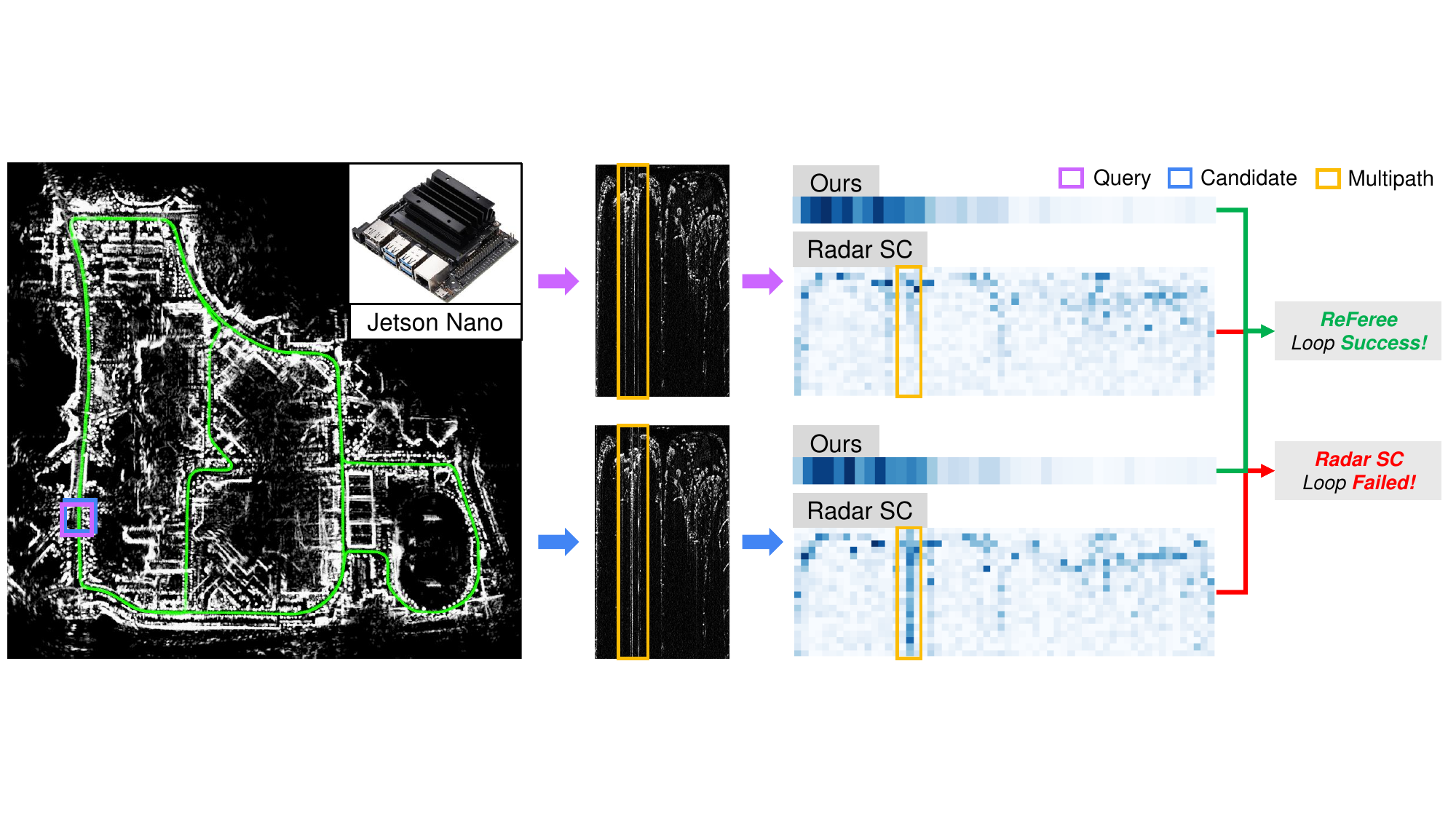}
    }
    \vspace{-0.5cm}
    \caption{The proposed place recognition successfully identifies the correct loop in radar images abundant with multipath and speckle noise. 
             In contrast, Radar SC \cite{kim2020mulran} exhibits vulnerability to multipath, often leading to the identification of incorrect loops.
             Also, 42$\times$1 shape of our lightweight descriptor enables us to perform the \ac{SLAM} with Nvidia Jetson Nano in the \textit{KAIST} sequence of \textit{Mulran} dataset \cite{kim2020mulran}.}	
\vspace{-0.5cm}
\label{fig:main}
\end{figure}

\bl{Unlike these methods, we propose a radar-based lightweight and robust global descriptor with a feature and free space called \textit{ReFeree} by using the radar image in polar coordinates without a cartesian converting process.}
The main contributions of this work are the following:
\begin{itemize}
    \item \textbf{Lightweight, Robust and Rotational Invariance Description:} 
    We achieve robust place recognition by a rapid feature extraction approach from the radar image and descriptor using free space information.
    As shown in \figref{fig:main}, the free space-based description can detect loops missed by the feature-based description since it attenuates the effects of multipath and speckle noise.
    Also, the proposed method can find loops in opposite directions via the rotational invariance property.
    In addition, the proposed descriptor that is at least 4$\times$ and up to 600$\times$ lighter compared to other methods and the KD-Tree searching process enhances usability on onboard computers.

    \item \textbf{Initial Heading and Semi-metric information:}     
    Unlike previous methods \cite{choi2024referee}, our approach enables semi-metric localization by estimating the 1-DoF heading between the revisited place and the current place.
    The estimated initial heading assists in making the registration performance efficient and robust, aiding in constructing a \ac{SLAM}.
        
    \item \textbf{Extensive Evaluation and Open-source:} 
    We evaluate the place recognition performance of the proposed method across various scenarios (i.e. single session, multi-session, and different weather conditions) including extreme environments.
    Also, we utilize our descriptor in Full-SLAM pipeline for practical validation and conduct it on NVIDIA Jetson Nano in online for lightweight validation and testing onboarding computational performance.
    Furthermore, we will contribute to the radar robotics community by sharing our proposed method as open source.
\end{itemize}

\section{Related Works}
Utilizing ranging sensors (e.g. radar, LiDAR) for place recognition has yielded remarkable results \cite{kim2018scan, kim2024narrowing, cai2022autoplace, meng2024mmplace, herraez2024spr, kim2020mulran, jang2023raplace, gadd2024open, choi2024referee}.
Recently, it has been studied in place recognition because, unlike LiDAR, it can provide information that is robust to adverse weather rather than clear weather.
Radar-based place recognition can be divided into two major categories; (i) single-on-chip (SoC) radar, which provides a very sparse point cloud along with velocity, and (ii) scanning radar, which can obtain dense images containing semi-metric information.
However, radar-based place recognition is still relatively in its infancy.
Thus, in this section, we discuss the LiDAR as well as the SoC radar and scanning radar research.

\subsection{Place Recognition for LiDAR}
\citet{he2016m2dp} who proposed M2DP analyzed the principal components of 3D local point clouds, which project them onto these principal components to create a 2D matrix and produce a vector using Singular Value Decomposition (SVD). 
Therefore, M2DP can represent space as a compact and efficient descriptor based on its principal component.
PointNetVLAD \cite{uy2018pointnetvlad} is a place recognition method, which proposes learning directly from 3D data through neural networks. 
Scan Context (SC) \cite{kim2018scan} is a 2D matrix descriptor that contains semi-metric information, where each matrix value is the highest value in the corresponding range and angle.
The initial heading can be estimated by semi-metric information and leads to successful registration by enhancing the performance of \ac{ICP}.
However, depending on the wavelength used by LiDAR, it is sensitive to ambient light and is especially vulnerable to adverse weather conditions.

\subsection{Place Recognition for SoC Radar} 
AutoPlace \cite{cai2022autoplace} aimed to improve place recognition robustness by exploiting the characteristics of the SoC radar. 
It utilized dynamic point removal, spatial-temporal feature embedding, and candidate refinement to enhance radar-based place recognition.
mmPlace \cite{meng2024mmplace} was a robust place recognition system that addresses common limitations of the SoC radar, such as the sparse point cloud and limited field of view (FOV).
%
It transformed intermediate-frequency signals into range azimuth heat maps and employed a spatial encoder for feature extraction. 
To tackle rotational and lateral variations, mmPlace used the rotating platform hardware to enhance the FOV, achieving high recall rates.
However, SoC radar necessarily utilizes more than one sensor for omnidirectional view due to limitations on the FOV, and it requires calibration between multiple SoC radars.

\subsection{Place Recognition for Scanning Radar}
Radar Scan Context (Radar SC) proposed with the \textit{Mulran} \cite{kim2020mulran} dataset is a 2D matrix descriptor similar to the SC \cite{kim2018scan} generated by directly resizing the radar image.
The algorithm exhibits rotational invariance, similar to Scan Context (SC), so it could recognize revisited locations in both forward and reverse driving directions.
However, Radar SC is too susceptible to speckle noise because it accepts noise as it is.
RaPlace \cite{jang2023raplace} utilizes the radon transform (RT) to create a sinogram and the Fast Fourier Transform (FFT) to generate rotational and translational invariant descriptors.
Despite achieving bidirectional invariance, the Fourier Transform-based descriptor has an issue that causes the loss of geometric information in metric space.
In \cite{gadd2024open}, RadarVLAD (RadVLAD) and FFT-RadarVLAD (FFT-RadVLAD) are proposed as a vector of locally aggregated descriptors that achieve rotational invariance using clustering of K-means.
FFT-RadVLAD is a descriptor that combines these algorithms to add the FFT for rotational invariance and time efficiency.
However, K-means clustering is vulnerable to speckle noise or multipath effect because it takes the image without pre-processing.
Also, it lost geometric information by utilizing FFT, which is similar to RaPlace.
\bl{Learning-based methodologies like \cite{gadd2020kidnapped} also exist, but they suffer from a similar drawback to \cite{gadd2024open}: they rely heavily on training data.}
\citet{hong2022radarslam} proposed a descriptor generation approach similar to M2DP \cite{he2016m2dp} as mentioned above.
The descriptor generation process in \cite{hong2022radarslam} entails dimensionality reduction, and it inevitably causes information loss.

We proposed a method that calculates the amount of free space to the farthest features for each angle-wise signal of a radar image in our previous work \cite{choi2024referee}.
However, since the descriptor is created for angle direction, it is vulnerable to rotational changes.
On the other hand, our methodology, based on scanning radar data, mitigates the effects of noise to ensure robustness and utilizes geometric information, resulting in additional performance gains in registration.
Unlike existing methodologies that utilize structural information, except for previous work, we utilize free space and show notable results in place recognition even in extreme environments.

\begin{figure*}[t]
	\centering
	\def\width{0.85\textwidth}%
    {%
        \includegraphics[clip, trim= 250 250 250 250, width=\width]{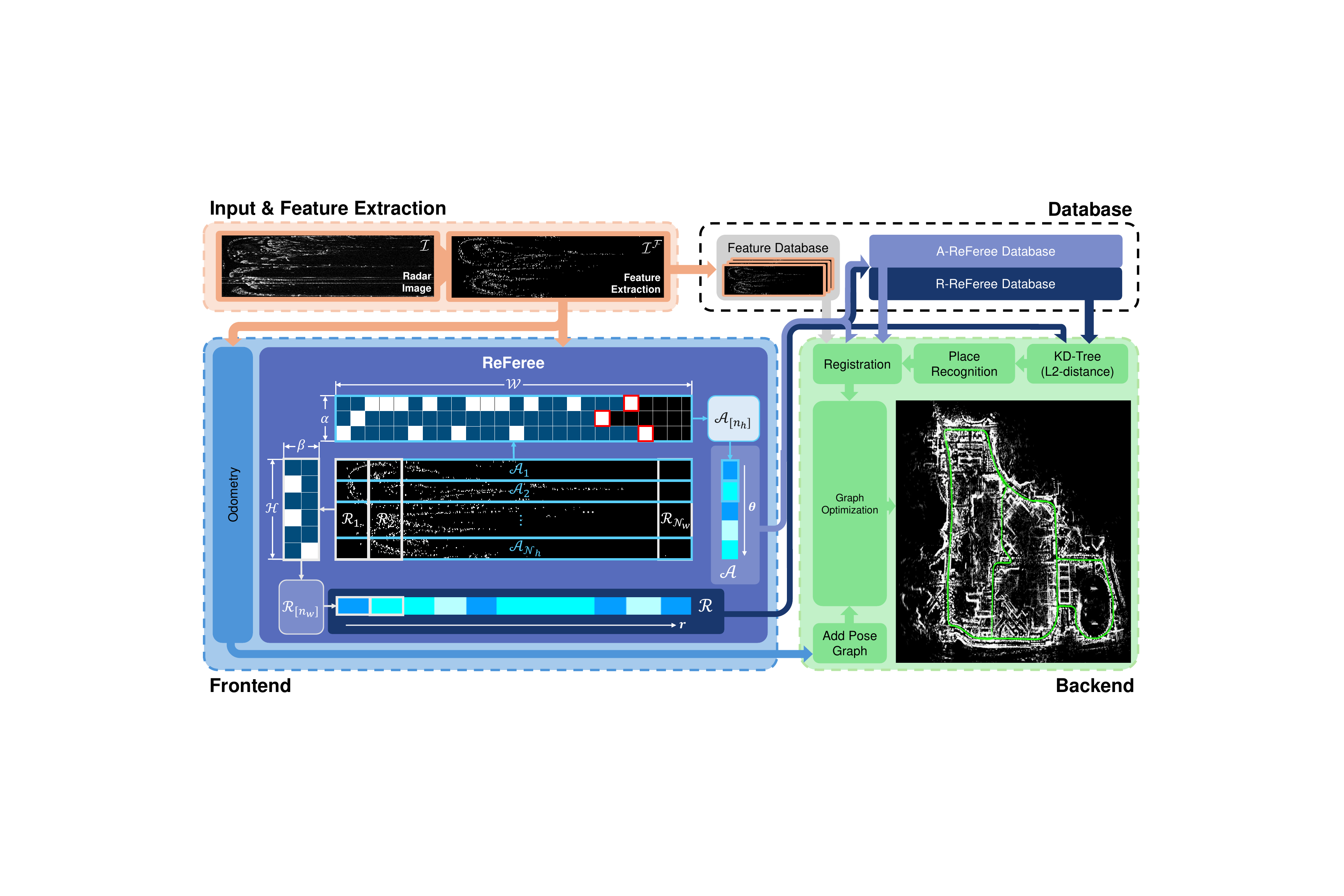}
    }
    \vspace{-0.3cm}
    \caption{
             Our method's pipeline.
             In the frontend, we perform feature extraction and generate descriptors: \textit{R-ReFeree} for place recognition and A-ReFeree for initial heading estimation.
             In the backend, we obtain a radar map through place retrieval and pose graph optimization.
             The white square and blue squares on the radar image in the \textit{ReFeree} module are range-wise block and angel-wise block respectively.
             And the white and blue squares that make up the range-wise block and angle-wise block represent feature and free space respectively, and the red square represents the farthest feature for a unit angle.} 
    \label{fig:pipeline}
    \vspace{-0.4cm}
\end{figure*}

\section{Method}
The intensity of radar image, which is the reflection coefficient of the target, is proportional to the Radar Cross Section (RCS) value which is the ratio of the reflected power to the incident power density.
The radar image that consists of an intensity defines $\mathcal{I} \in\mathbb{R}^{\mathcal{H} \times \mathcal{W}}$ in polar coordinates, where $\mathcal{H}$ is the height of the image that is associated with angle and $\mathcal{W}$ is the image width which is associated with range.
As reported by \cite{harlow2023new}, a variety of noise characteristics across the sensor range, including spurious returns, complex speckle noise, and multipath reflections where multiple signals are returned from the same object, have high-intensity values, which can mask meaningful information.
In this section, for mitigating the effects of these noises, we describe the process of two kinds of \textit{ReFeree} descriptor from the radar image $\mathcal{I}$, \textit{R-ReFeree} for place recognition and \textit{A-ReFeree} for estimating initial heading.
In addition, our method integrates odometry and graph optimization for SLAM, and the results will be discussed in Section~\uppercase\expandafter{\romannumeral4}-G.

\subsection{Feature Extraction}
To extract meaningful information from the noisy radar image, we use a feature extraction algorithm inspired by \cite{cen2018precise}.
This algorithm involves decomposing the signal into high and low-frequency signals and integrating them according to a nonlinear Gaussian filter.
Finally, the integrated signal represents the power return at each range bin and goes through the thresholding process to extract valid features from the raw radar signal.
Unlike this algorithm, we adopt a simple and linear uniform filter in the signal smoothing step for faster computation, making it more suitable for onboard computers.
These features extracted from the integrated signal constitute the radar feature image $\mathcal{I}^{\mathcal{F}}$.
\subsection{Place Description (ReFeree)}
Although the feature extraction process solves the aforementioned problem that the high-intensity value masks meaningful information, the noise cannot be removed completely.
To relax the impact of false detection including noise, we utilize free space to generate the descriptor instead of features.
To describe the detail characteristic of the free space we utilized, we split the result of the feature extraction process into true detection which classifies the feature successfully, and false detection which fails to classify the feature correctly.
False detection makes it challenging to estimate the position of objects accurately, and consequently, it critically impacts methodologies that rely on features alone to generate descriptors.
So, we concentrate on the point that false detection of feature extraction consists of detecting free space as features and vice versa.
These two false detection cases can be analyzed from the perspective of feature and free space, as shown in \tabref{tab:false}.
\begin{table}[h]
    \vspace{-0.3cm}
    \caption{Constitues of False Detection}
    \centering\resizebox{0.4\textwidth}{!}{
            \begin{tabular}{c|cc}
                \toprule \hline
                                          & \multicolumn{2}{c}{Perspective}        \\ \hline
                Ground Truth / Prediction & Feature                   & Free space \\ \hline
                Feature / Free space      & ${FN}^{\mathcal{F}}$      & ${FP}^{f}$ \\ \hline
                Free space / Feature      & ${FP}^{\mathcal{F}}$      & ${FN}^{f}$ \\ \hline \bottomrule
            \end{tabular}
        }
    \label{tab:false}
    \vspace{-0.15cm}
\end{table}
Detecting free space as a feature is a false positive from the perspective of feature extraction, but it is a false negative from the perspective of extracting free space.
Conversely, detecting a feature as free space is a false negative from the perspective of feature extraction, but it is a false positive from the perspective of extracting free space.
This relationship can be formulated as follows:
\begin{equation}
    \#False = \#{FP}^{\mathcal{F}}+\#{FN}^{\mathcal{F}} = \#{FN}^{f}+\#{FP}^{f},
    \label{equ:false}
\end{equation}
\noindent where superscript $\mathcal{F}$ and $f$ are feature and free space respectively, and the notation $\#$ means the number of.
This reciprocal relationship suggests that the number of false detections from the perspective of free space is equal to the number of false detections interpreted from the perspective of features.
The impact of false detection on the description can be numerically calculated by the false detection rate, which is the number of false detections ($\#False$) divided by the number of features ($\#Feature$) or the amount of free space ($\#Free\;space$).
\begin{table}[h]
    \vspace{-0.25cm}
    \caption{Mean Feature and Free space of Mulran Sequences}
    \centering\resizebox{0.3\textwidth}{!}{
            \small
            \begin{tabular}{c|c|c}
                \toprule \hline
                              & DCC 01     & KAIST 03 \\ \hline
                \# Free space & 1,340,891  & 1,340,684 \\ \hline
                \# Feature    & 3,108      & 3,315 \\ \hline \bottomrule
            \end{tabular}
        }
    \label{tab:numfeaturefree}
\end{table}

As shown in \tabref{tab:numfeaturefree}, the number of free spaces is significantly larger than the number of features, which means that the false detection rate of free spaces is relatively lower than the false detection rate of features in a single radar image.
This contributes significantly to mitigating the impact of misclassification.

\subsubsection{R-ReFeree}
Unlike previous work \cite{choi2024referee} that fails to achieve rotational invariance, making it vulnerable to angular changes, we obtain rotational invariance by using range-wise blocks as shown in \figref{fig:pipeline}.
First, we denote free space identifier as:
\begin{equation}
    \psi(\mathrm{p}) = 
    \begin{cases}
        1, & \mbox{if }\mathrm{p}\mbox{ is a \textit{free-space}} \\
        0, & \mbox{otherwise},
    \end{cases}
    \label{equ:free or not}
\end{equation}
where $\mathrm{p}$ is a pixel value of radar feature image $\mathcal{I}^{\mathcal{F}}$.
Second, we denote the descriptor, the number of free space, $\mathcal{R}$ utilizing the free space identifier $\psi$ as follows:
\begin{equation}
	\label{equ:r_referee}
    \mathcal{R}_{[\mathbf{n}_{w}]} = \sum_{j=\beta[\mathbf{n}_{w}]}^{\beta([\mathbf{n}_{w}]+1)} \left( \sum_{i=1}^{\mathcal{H}} \psi(\mathcal{I}^{\mathcal{F}}_{ij}) \right), \quad \mathcal{R} \in\mathbb{R}^{\mathcal{N}_{w}},
\end{equation}
\noindent where $\beta$ is the size of the range-wise block.
The $\mathbf{n}_{w}=[1, 2, ..., \mathcal{N}_{w}]$, where $\mathcal{N}_{w}=\frac{\mathcal{W}}{\beta}$, is descriptor index vectors and $[\mathbf{n}_{w}]$ means each element of $\mathbf{n}_{w}$.

\subsubsection{A-ReFeree}
The \textit{A-ReFeree}, as a 1-DoF heading estimator is introduced as follows:
\begin{equation}
    \label{equ:hat_a_referee}
    \mathcal{A}_{[\mathbf{n}_{h}]} = \sum_{i=\alpha[\mathbf{n}_{h}]}^{\alpha([\mathbf{n}_{h}]+1)}	\left( \sum_{j=1}^{r_i} \psi(\mathcal{I}^{\mathcal{F}}_{ij}) \right), \quad \mathcal{A} \in\mathbb{R}^{\mathcal{N}_{h}},
\end{equation}
\noindent where $\alpha$ is the size of the angle-wise block.
The $\mathbf{n}_{h}=[1, 2, ..., \mathcal{N}_{h}]$, where $\mathcal{N}_{h}=\frac{\mathcal{H}}{\alpha}$, is descriptor index vectors and $[\mathbf{n}_{h}]$ means each element of $\mathbf{n}_{h}$.
The notation $r_i$ is element of farthest feature index vector $\bold{R}=[r_1,\cdots,r_\mathcal{H}]$, and it is matched with each row of $\mathcal{I}^{\mathcal{F}}$.
When configuring \textit{A-ReFeree} as shown in \figref{fig:pipeline}, we count free space up to the farthest feature due to radar's ability to penetrate particles.

\textit{A-ReFeree} can recognize the revisited place where the angle is different from the current place through the angle-shifting module, but it must consume an amount of time.
Thus, we only utilize the \textit{A-ReFeree} as the 1-DoF heading estimator for detected loops.
As a result, by properly utilizing two \textit{ReFerees}, we achieve rotational invariance and initial estimation simultaneously.
This efficient division of labor also enables real-time place recognition from onboard radar sensor measurements.

\subsection{Place Recognition}
We create the \textit{R-ReFeree} and recognize the revisited places to correct the accumulated pose errors. 
We construct a KD-Tree from the candidates to facilitate faster searching, and the distance between the query \textit{R}-\textit{ReFeree} $\mathcal{R}^q$ and the candidate \textit{R}-\textit{ReFeree} $\mathcal{R}^c$ is calculated using the L2-norm as follows:
\begin{equation}
    \sigma(\mathcal{R}^q, \mathcal{R}^c) = {||\mathcal{R}^q - \mathcal{R}^c||_{2}} .
    \label{equation:initial_index}
\end{equation}
The optimal candidate $c^*$ for query descriptor $\mathcal{R}^q$ can be formulated as:
\begin{equation}
	\label{equ:place_retrieval}
    c^{*} = \underset{c \in \mathcal{C}}{\argmin}\;\sigma(\mathcal{R}^q, \mathcal{R}^c),\;\;s.t.\;\;\sigma(\mathcal{R}^q, \mathcal{R}^c) < \tau,
\end{equation}
where the $\mathcal{C}$ is place candidates set, $\tau$ is distance threshold.

\subsection{SLAM}
\subsubsection{Registration}
We can recognize the revisited places through the \textit{R}-\textit{ReFeree}, and obtain the initial heading of the two places through the \textit{A}-\textit{ReFeree}.
Due to the semi-metric information of the \textit{A}-\textit{ReFeree}, shifting candidates implies that the index of the closest distance represents $\hat{n}$ as a 1-DoF heading following:
\begin{equation}
    \hat{n} = \argmin\limits_{n \in \mathbf{n}_{h}}({d(\mathcal{A}^q, \mathcal{A}}^{c^*}_{n})) ,
    \label{equation:initial_index}
\end{equation}
where the function $d(\cdot,\cdot)$ is cosine distance.
We convert the estimated initial heading into a rotation matrix at loop $\mathcal{L}$:
\begin{equation}
 T^{\mathcal{L}}_{(q, \hat{q})} = \begin{bmatrix}
                  cos\left( \hat{n} \times \frac{360}{\mathcal{N}_{h}} \right) & -sin\left( \hat{n} \times \frac{360}{\mathcal{N}_{h}} \right) \\ 
                  sin\left( \hat{n} \times \frac{360}{\mathcal{N}_{h}} \right) &  cos\left( \hat{n} \times \frac{360}{\mathcal{N}_{h}} \right)
              \end{bmatrix} .	
\end{equation}
The initial aligned query point cloud $\hat{\mathcal{P}}^{q}$ for scan-to-scan matching is calculated from query point cloud $\mathcal{P}^{q}$ corresponding with $\mathcal{A}^q$ and $T^{\mathcal{L}}_{(q,\hat{q})}$ as follows:
\begin{equation}
\hat{\mathcal{P}}^{q} = {T^{\mathcal{L}}_{(q,\hat{q})}}^{-1} \mathcal{P}^{q} .
\end{equation}
We perform scan-to-scan matching between the initial aligned query point cloud $\hat{\mathcal{P}}^{q}$ and the candidate point cloud $\mathcal{P}^{c}$ to compute the relative transform.
Scan-to-scan matching algorithm $\xi$ can be formulated as:
\begin{equation}
T^{\mathcal{L}}_{(\hat{q},c^*)} = \xi(\hat{\mathcal{P}}^{q},\mathcal{P}^{c^*}),
\end{equation}
where $T^{\mathcal{L}}_{(\hat{q},c^*)}$ is relative transform between $\hat{\mathcal{P}}^{q}$ and $\mathcal{P}^{c^*}$.
We adopt the $\xi$ as Nano-GICP \cite{chen2022direct}, which combines the fast speed registration and the efficient construction of KD-Tree using the open-source library NanoFLANN.
Finally, we obtain an optimal transformation matrix as follows:
\begin{equation}
T^{\mathcal{L}}_{(q,c^*)} = T^{\mathcal{L}}_{(q, \hat{q})} T^{\mathcal{L}}_{(\hat{q},c^*)}.
\end{equation}
To prevent mismatches between query place and candidate place, we progress a geometry verification process by calculating the fitness score through the \ac{ICP} result, which enables us to validate the accurate loop once more.

\subsubsection{Pose Graph Optimization}
As pose graph $\mathbf{X}$ from radar odometry \cite{burnett_ral21} contains its accumulated errors due to sensor measurement, so loop closure is essential to build a consistent map.
The pose graph optimization of the map is represented as:
\begin{equation}
\begin{aligned}
\hat{\mathbf{X}}=\underset{\mathbf{X}}{\operatorname{argmin}} & \sum \varphi(\mathbf{X}) .
\end{aligned}
\end{equation}
Also, pose graph optimization function $\varphi(\cdot)$ is defined as:
\begin{equation}
\begin{aligned}
\varphi(\mathbf{X})= & \sum_t\left\|f\left(\mathbf{x}_t, \mathbf{x}_{t+1}\right)- \zeta({\mathbf{z}}_{t}, {\mathbf{z}}_{t+1})\right\|_{\sum_t}^2 \\
& +\sum_{\langle q, c^* \rangle \in \mathbb{L}} \left\|f\left(\mathbf{x}_q, \mathbf{x}_{c^*}\right)-\xi(\hat{\mathcal{P}}^{q},\mathcal{P}^{c^*}) \right\|_{\sum_{q, c^*}}^2,
\end{aligned}
\end{equation}
where $f(\cdot, \cdot)$ is 3-DoF relative pose estimator, $\zeta(\cdot, \cdot)$ is radar odometry estimator with features $\mathbf{z}_{t}$ that contains the timestamp, $\xi(\cdot, \cdot)$ is scan-to-scan matching between $\hat{\mathcal{P}}^{q}$ and $\mathcal{P}^{c^*}$ captured in $\mathcal{L} \in \mathbb{L}$, and $\sum_t$, $\sum_{q, c^*}$ are uncertainty for an odometry factor and a loop closure factor, respectively.
\section{Experiment Results}
All experiments were conducted on desktop AMD Ryzen 7 5700X with RTX 3080 to evaluate overall performances and Nvidia Jetson Nano to assess onboard computing capabilities of the \ac{SLAM} pipeline. We compared the proposed method to the recent open-source approaches. The section \uppercase\expandafter{\romannumeral5}-B provides a detailed discussion of the evaluation metrics.
Also, we differentiated experiment results for easy comprehension: \textbf{first rank} in bold and the \underline{second rank} in underline.

\subsection{Datasets}
For exhaustive validation of the proposed method, we selected three kinds of datasets that installed different radar models as shown in \tabref{tab:datasets}.
In addition, we compared our descriptor with other descriptors in all datasets under the place density of 20m.
More detailed datasets are given below.
\begin{table}[h]
\caption{Datasets Comparision}
\centering\resizebox{0.49\textwidth}{!}{
\begin{tabular}{c|c|c|c|c}
\toprule
       & \textbf{Radar Model}       & \textbf{Session} & \textbf{Complexity}                         & \textbf{Weather Diversity}       \\ \hline
Mulran & Navtech CIR204-H           & Single           & $\bigstar$                                  & $\bigstar$                       \\ \hline
Oxford & Navtech CTS350-X           & Multi            & $\bigstar$                                  & $\bigstar$                       \\ \hline
OORD   & Navtech CTS350-X           & Multi            & $\bigstar$ $\bigstar$ $\bigstar$            & $\bigstar$ $\bigstar$            \\ \hline
Boreas & Navtech CIR304-H           & Multi            & $\bigstar$                                  & $\bigstar$ $\bigstar$ $\bigstar$ \\ \bottomrule
\end{tabular}}
\label{tab:datasets}
\end{table}

\subsubsection{Mulran \cite{kim2020mulran}}
We used \textit{DCC} and \textit{KAIST} sequences for evaluating the performance of single-session place recognition.
The \textit{DCC} sequence offers structural diversity, and the \textit{KAIST} sequence enables testing of various dynamic objects.
Both selected sequences have lane-level transformations for robustness evaluation, notably the presence of a reverse loop in \textit{DCC}.
Since this dataset contains appropriate loops to correct for the accumulated error in odometry, we attached the \ac{SLAM} results.

\subsubsection{OORD \cite{gadd2024oord}}
This dataset is an outdoor environment that contains unpaved terrain and sloped trails with diverse weather conditions, such as snowfall.
All sequences have more than one operation across several periods, so it is appropriate to utilize them for multi-session evaluation.
\textit{Bellmouth} sequence captures offroad environments such as gravel tracks.
\textit{Hydro} sequence presents a snow-covered route along the mountain river.
\textit{Maree} sequence contains multiple snow-covered sloped trails and driving data with total darkness.

\subsubsection{Oxford Radar RobotCar \cite{barnes2020oxford}}
\bl{The \textit{Oxford Radar RobotCar} dataset aims to evaluate large-scale urban autonomy in a structured environment.
Since it has relatively fewer loops than Mulran, we used it to validate the results of the multi-session evaluation.}

\subsubsection{Boreas \cite{burnett2023boreas}}
\textit{Boreas} dataset is collected by a repeat route for one year with seasonal variations.
Given these characteristics, \textit{Boreas} dataset is suitable to evaluate the long-term weather-robust place recognition performance.

\subsection{Evaluation Metrics}
\subsubsection{PR curve and AUC score}
The Precision-Recall (PR) curve is utilized to assess the overall performance of place recognition. 
Precision and recall are defined as:
\begin{equation}
\text { Precision }=\frac{\mathrm{TP}}{\mathrm{TP}+\mathrm{FP}}, \quad \text { Recall }=\frac{\mathrm{TP}}{\mathrm{TP}+\mathrm{FN}},
\end{equation}
where TP, FP, and FN are true positive, false positive, and false negative, respectively. 
We also determined the AUC (Area Under the Curve) score that represents the area under the PR curve.

\subsubsection{F1-Recall curve}
The F1-Recall curve is used to check how the TP is maintained despite the large number of loops.
F1 score is a harmonic mean of precision and recall as:
\begin{equation}
\text { F1 score }=\frac{2 \times \mathrm{Precision} \times \mathrm{Recall}}{\mathrm{Precision}+\mathrm{Recall}}.
\end{equation}

\subsubsection{Recall@1}
We employed Recall@1 as a metric for performance in finding loops, rather than the PR curve. 
\begin{equation}
\text { Recall@1 }=\frac{\mathrm{TP}}{\mathrm{GT}},
\end{equation}
where GT is the number of ground truth. 

\subsubsection{Processing Time}
Loop detection can be divided into two processes: creating descriptors and finding revisited places. 
Therefore, we measured and compared the average speeds of these two processes for analysis.

\subsubsection{Rotation Error}
We evaluated a Rotation Error (RE), which is the error between the estimated rotation and the ground truth rotation calculated by,
\begin{equation}
\text { RE }= || \hat{\alpha} - \alpha ||,
\end{equation}
where $\hat{\alpha}$ is the estimated rotation angle and $\alpha$ is the ground truth rotation angle.

\subsubsection{APE}
We used the APE (Absolute Pose Error) metric to verify the efficiency of the pose graph optimization as follows:
\begin{equation}
\text {APE} = \sqrt{\frac{1}{N}\sum_{i=1}^{N}||p_{i, est} - p_{i, gt}||},
\end{equation}
where $p_{i, est}$ means the optimized pose, $p_{i, gt}$ represents the ground truth pose, and $N$ is the total number of poses.

\subsection{Performance on Description}
In this section, we introduced two properties that make our descriptor functionally superior to other descriptors.
\begin{table}[h]
\vspace{-0.3cm}
\caption{Descriptor Comparision}
\centering\resizebox{0.49\textwidth}{!}{
\begin{tabular}{l|c|c|c|c}
\toprule
\textbf{Approach}          & \textbf{Size [B]} $\downarrow$ & \textbf{Shape} $\downarrow$ & \textbf{Semi-Metric} & \textbf{Initial Heading}  \\ \hline
\underline{Radar SC}       & \underline{1,616}               & 20$\times$60                & \checkmark  & $\times$             \\
RaPlace                    & 3,008                           & \underline{20$\times$18}    & $\times$    & $\times$             \\
RadVLAD                 & 262,272                        & 32768$\times$1              & $\times$    & $\times$             \\
FFT-RadVLAD              & 262,272                        & 32768$\times$1              & $\times$    & $\times$             \\
\textbf{ReFeree (Ours)}    & \textbf{464}                   & \textbf{42$\times$1}        & \checkmark  & \checkmark           \\ \bottomrule
\end{tabular}}
\label{tab:comparision_desc}
\end{table}
\subsubsection{Lightness} 
We reported the lightness of the descriptors. 
As shown in \tabref{tab:comparision_desc}, the VLAD-based methodology is approximately 600$\times$ larger than ours, while RaPlace is roughly 6$\times$ larger.
Also, our lightweight descriptors combined with KD-Tree search can perform place recognition in online on the onboard computer.
\subsubsection{Semi-Metric and Initial Heading} 
Only Radar SC and our method contain semi-metric information, but Radar SC is not given an initial heading.
Therefore, only the proposed method can help the performance of ICP with an initial heading, which we will discuss in Section \uppercase\expandafter{\romannumeral4}-D-2.


\subsection{Place Recognition in Single session}
\subsubsection{Overall performance and Processing Time} 
We evaluated various metrics on the \textit{Mulran} dataset.
\tabref{tab:single} illustrates the overall place recognition performance. 
We observed that the proposed method outperforms all metrics in all sequences.
While it is interesting that VLAD-based descriptors have a fast detection speed relative to their size, the drawbacks are that they require pre-training, and the speed decreases proportionally with longer trajectories.
\begin{table}[h]
\caption{Overall Performance Evaluation in single session}
\centering\resizebox{0.49\textwidth}{!}{
\begin{tabular}{c|cccccc}
\toprule
                          & \multicolumn{5}{c}{\textbf{DCC 01}}                                                \\ \hline
\textbf{Approach}          & \textbf{Recall@1 $\uparrow$}    & \textbf{F1 max $\uparrow$}       & \textbf{AUC $\uparrow$}      & \textbf{Proc. Time [s] $\downarrow$} & \textbf{RE ($^\circ$)}\\ \hline
Radar SC                   & 0.683                & \underline{0.852}     & 0.908             &  0.368 & - \\
\underline{RaPlace}        & 0.641                & 0.849                 & \underline{0.917} &  0.359 & - \\
RadVLAD                 & 0.507                & 0.819                 & 0.885             &  \underline{0.081}& - \\
FFT-RadVLAD              & \underline{0.706}    & 0.779                 & 0.874             &  0.088 & - \\
\textbf{ReFeree (Ours)}    & \textbf{0.723}       & \textbf{0.861}        & \textbf{0.936}    &  \textbf{0.069} & \textbf{23.32} \\ \hline \hline
                          & \multicolumn{5}{c}{\textbf{KAIST 03}}                                                \\ \hline
\textbf{Approach}          & \textbf{Recall@1 $\uparrow$}    & \textbf{F1 max $\uparrow$}       & \textbf{AUC $\uparrow$}      & \textbf{Proc. Time [s] $\downarrow$} & \textbf{RE ($^\circ$)}\\ \hline
\underline{Radar SC}       & 0.896                & \underline{0.966}     & \underline{0.991}    &  0.375 & - \\
RaPlace                    & 0.932                & 0.938                 & 0.984             &  0.383 & - \\
RadVLAD                 & 0.833                & 0.931                 & 0.978             &  \underline{0.102} & - \\
FFT-RadVLAD              & \underline{0.948}    & 0.937                 & 0.985             &  0.108 & - \\
\textbf{ReFeree (Ours)}    & \textbf{0.955}       & \textbf{0.970}        & \textbf{0.992}    &  \textbf{0.070} & \textbf{2.14}\\ 
\bottomrule
\end{tabular}}
\label{tab:single}
\end{table}

\subsubsection{Rotation Error and Initial Heading Estimation}
\begin{figure}[h]
	\centering
	\def\width{0.49\textwidth}%
    {%
        \includegraphics[clip, trim= 0 130 0 160, width=\width]{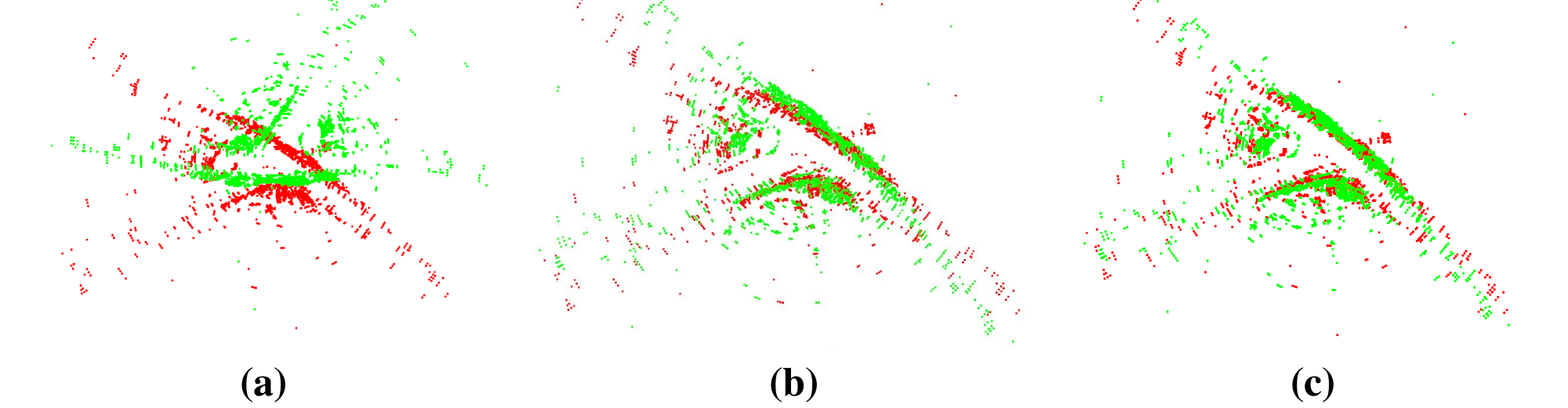}
    }
        \vspace{-0.6cm}
        \caption{Query point cloud (red) and candidate point cloud (green) at frame $1492^{nd}$ and $1041^{st}$ in DCC 01, respectively.
                 (a) The point clouds without estimating the initial heading. 
                 (b) After transforming the source point cloud with the initial heading.
                 (c) The visualization after optimization with ICP in the state of (b).}
\label{fig:reverse_loop}
\vspace{-0.4cm}
\end{figure}

Initial heading plays a pivotal role in the \ac{SLAM} pipeline as it dramatically enhances ICP performance.
We compared the RE calculated by our method when the F1 score is maximum.
As shown in \tabref{tab:single}, our method presents the mean RE in both sequences.
Our method can also estimate the initial heading between frames $1492^{nd}$ and $1041^{st}$, which is the opposite directional relationship for the reverse loop.
However, other methods cannot estimate the initial rotation because they do not preserve rotational information.
\figref{fig:reverse_loop}(b) represents that our method can estimate the initial heading and \figref{fig:reverse_loop}(c) shows successful alignment results.

\begin{figure}[h]
         \centering
        \def\width{0.49\textwidth}%
    {%
        \includegraphics[clip, trim= 0 0 0 0, width=\width]{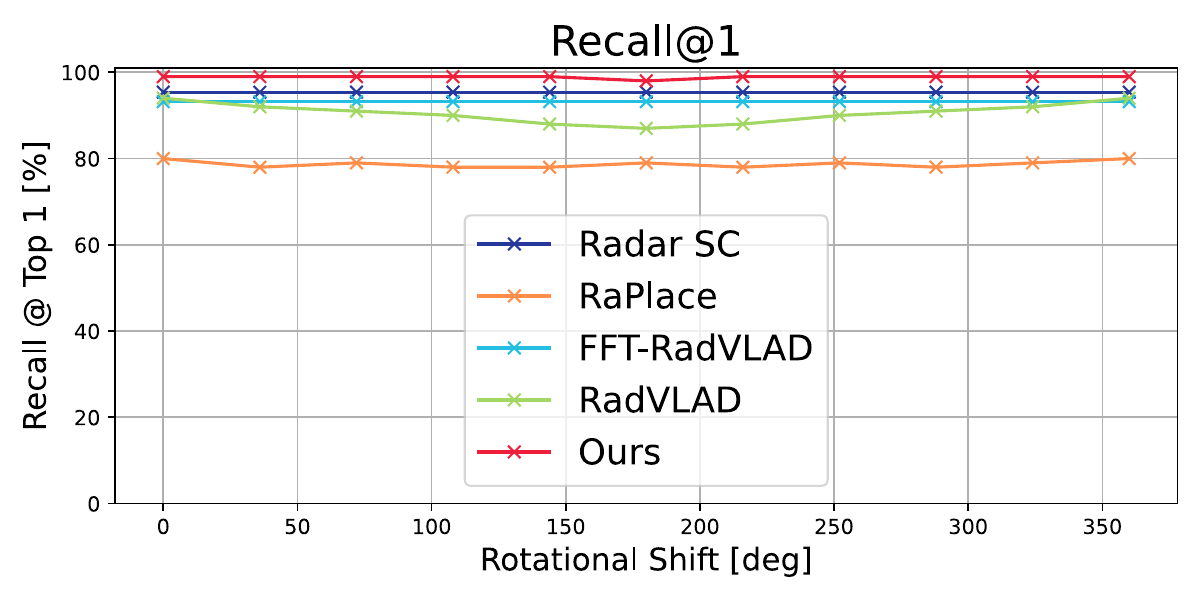}
    }
        \vspace{-0.6cm}
        \caption{Recall@1 obtained by shifting the columns of the query radar image in the Hydro sequence. The radar image in the Hydro sequence implies that a shift of one column is about 0.9 degrees.}
\label{fig:rot_inv}
\vspace{-0.5cm}
\end{figure}

\subsection{Place Recognition in Multi session}
\subsubsection{Rotational Invariance} 
Since radar images contain semi-metric information, shifting one column means a change in angle.
Therefore, we analyzed how consistent the candidate radar images can be found when shifting the column of the query radar images to check for rotational invariance.
\figref{fig:rot_inv} demonstrate the rotational invariance of our method.
Only Radar SC, FFT-RadVLAD, and our method maintain consistent performance. 
We can check tiny variations in our method because the region of the kernel is slightly different in the feature extraction step when shifting.
Also, RadVLAD, which cannot get the rotational invariance via FFT, does not perform consistently, and RaPlace also has a ray processing of radon transform that interferes with consistency.

\subsubsection{PR curve and F1-Recall curve} 
\begin{figure}[h]
    \vspace{-0.3cm}
    \centering
    \def\width{0.47\textwidth}
    \includegraphics[clip, trim= 220 0 240 10, width=\width]{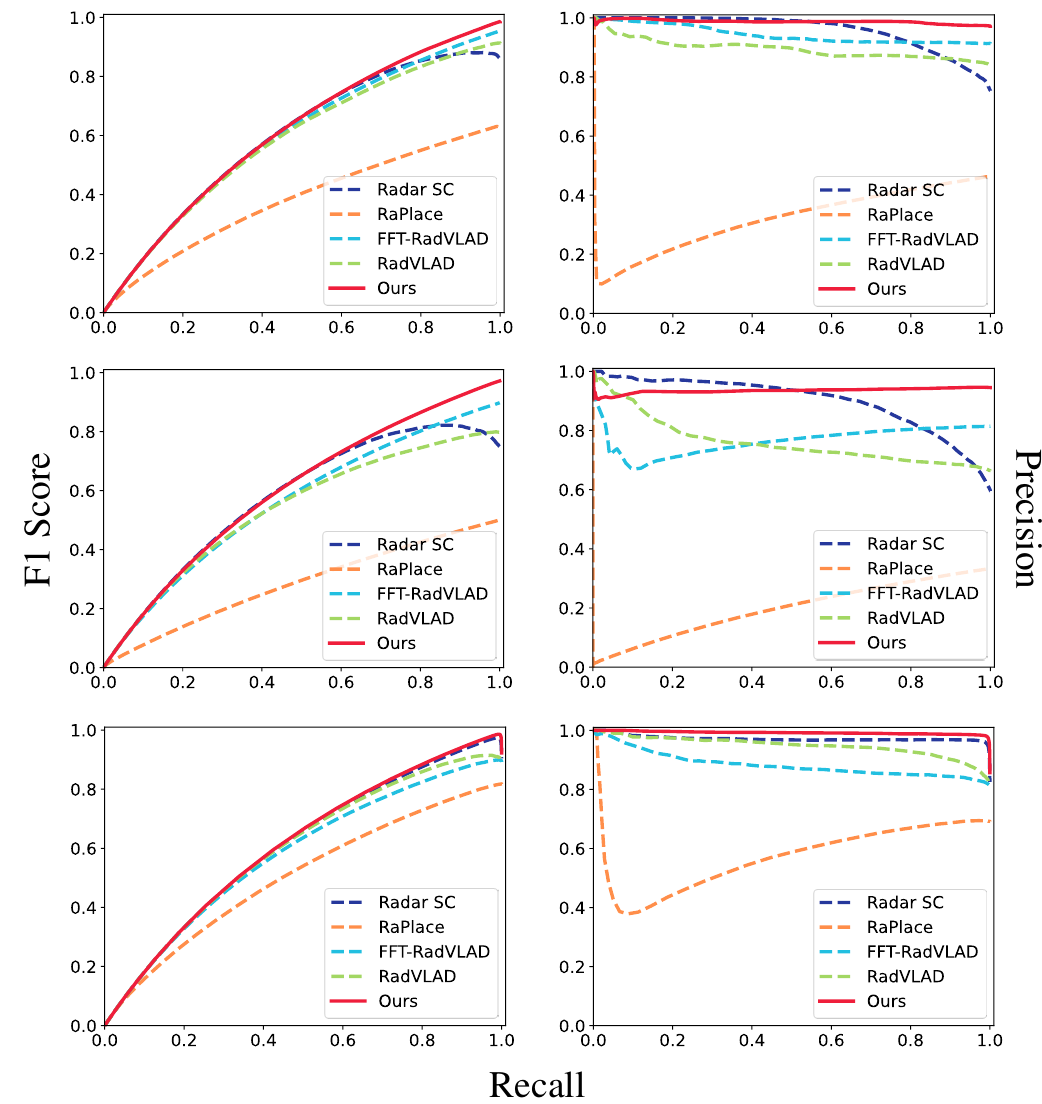}
    \vspace{-0.2cm}
    \caption{Top, middle, and bottom are \textit{Bellmouth}, \textit{Maree}, and \textit{Hydro}'s F1-Recall and PR curve in order from the left.}
    \label{fig:oord_pr}
\end{figure}

In extreme environments, the sparsity of features from structural information makes accurate loop validation through geometry verification challenging.
The PR curve in \figref{fig:oord_pr} indicates that the proposed method maintains high precision as recall increases, implying that it effectively distinguishes true loops.
\begin{table}[h]
\caption{Overall Performance Evaluation in Extreme Environment}
\centering\resizebox{0.49\textwidth}{!}{
\begin{tabular}{c|l|cccc}
\toprule
\textbf{Sequence}          & \textbf{Approach}          & \textbf{Recall@1 $\uparrow$}    & \textbf{F1 max $\uparrow$}       & \textbf{AUC $\uparrow$}      & \textbf{Proc. Time [s] $\downarrow$} \\ \hline
\multirow{5}{*}{Bellmouth} & Radar SC                   & 0.772                & 0.893                 & \underline{0.964}  &  0.394 \\
                           & RaPlace                    & 0.023                & 0.099                 & 0.040              &  0.628 \\
                           & \underline{RadVLAD}     & \underline{0.839}    & \underline{0.909}     & 0.881              &  \underline{0.386} \\
                           & FFT-RadVLAD              & 0.024                & 0.127                 & 0.064              &  0.390 \\
                           & \textbf{ReFeree (Ours)}    & \textbf{0.979}       & \textbf{0.986}        & \textbf{0.987}     &  \textbf{0.142} \\ \hline
\multirow{5}{*}{Maree}     & Radar SC                   & 0.634                & 0.827                 & \underline{0.902}  &  \underline{0.393} \\
                           & RaPlace                    & 0.351                & 0.500                 & 0.198              &  0.798 \\
                           & RadVLAD                 & 0.670                & 0.779                 & 0.727              &  0.542 \\
                           & \underline{FFT-RadVLAD}  & \underline{0.854}    & \underline{0.897}     & 0.765              &  0.547 \\
                           & \textbf{ReFeree (Ours)}    & \textbf{0.993}       & \textbf{0.971}        & \textbf{0.935}     &  \textbf{0.152} \\ \hline
\multirow{5}{*}{Hydro}     & \underline{Radar SC}       & \underline{0.954}    & \underline{0.972}     & \underline{0.970}  &  \underline{0.394} \\
                           & RaPlace                    & 0.799                & 0.818                 & 0.578              &  0.710 \\
                           & RadVLAD                 & 0.940                & 0.911                 & 0.943              &  0.450 \\
                           & FFT-RadVLAD              & 0.932                & 0.899                 & 0.882              &  0.452 \\
                           & \textbf{ReFeree (Ours)}    & \textbf{0.990}       & \textbf{0.986}        & \textbf{0.992}     &  \textbf{0.148} \\ 
\bottomrule
\end{tabular}}
\label{tab:oord_overall}
\vspace{-0.3cm}
\end{table}
\noindent  The F1 recall curve in \figref{fig:oord_pr} and \tabref{tab:oord_overall} demonstrates the ability to distinguish loops and indicates that our descriptor detects many loops. 

\subsubsection{Overall performance} 
To evaluate the performance of free space in extreme environments, we utilized the \textit{OORD} dataset. 
Unlike other methods that rely on structure, the proposed method achieves remarkable results.
In particular, the processing time, which is similar to RadVLAD and FFT-RadVLAD in the previous section, is almost 3$\times$ faster.
\tabref{tab:oord_overall} shows that free space can robustly describe places in extreme environments where features are scarce.

In \tabref{tab:oxford}, we also validated in the structured environment called \textit{Oxford Radar Robotcar} Datasets. 
Since the distinct structural information also implies distinct free space information, our method outperforms similar to the results of Section \uppercase\expandafter{\romannumeral4}-D.

\begin{table}[h]
\vspace{-0.3cm}
\caption{Overall Performance Evaluation in Oxford Radar RobotCar}
\centering\resizebox{0.49\textwidth}{!}{
\begin{tabular}{c|l|cccc}
\toprule
\textbf{Sequence}                   & \textbf{Approach}          & \textbf{Recall@1 $\uparrow$}    & \textbf{F1 max $\uparrow$}       & \textbf{AUC $\uparrow$}      & \textbf{Proc. Time [s] $\downarrow$} \\ \hline
                                    & Radar SC                   & 0.655                & 0.887                 & \textbf{0.960}    &  0.391 \\
           2019-01-10               & RaPlace                    & 0.666                & 0.799                 & 0.783             &  0.651 \\
                to                  & \underline{RadVLAD}        & \textbf{0.813}       & \textbf{0.897}        & 0.949             &  \underline{0.358} \\
           2019-01-18               & FFT-RadVLAD                & 0.666                & 0.800                 & 0.849             &  0.364 \\
                                    & \textbf{ReFeree (Ours)}    & \underline{0.803}    & \textbf{0.897}        & \underline{0.951} &  \textbf{0.125} \\ \hline
                                    
                                    & Radar SC                   & 0.348                & 0.701                 & 0.812             &  0.392 \\
           2019-01-16               & RaPlace                    & 0.437                & 0.608                 & 0.545             &  0.664 \\
               to                   & \textbf{RadVLAD}           & \textbf{0.693}       & \textbf{0.821}        & \underline{0.859} &  \underline{0.351} \\
           2019-01-18               & FFT-RadVLAD                & 0.427                & 0.605                 & 0.588             &  0.355 \\
                                    & \textbf{ReFeree (Ours)}    & \underline{0.677}    & \underline{0.816}     & \textbf{0.875}    &  \textbf{0.144} \\  \bottomrule

\end{tabular}}
\vspace{-0.4cm}
\label{tab:oxford}
\end{table}

\subsection{Place Recognition in Diverse Weather Condition}
\subsubsection{Overall performance} 
Although radar sensor data is robust in harsh weather conditions compared to the camera or LiDAR, it still generates noisy and harsh signals.
\tabref{tab:boreas_overall} shows the overall performance of the robustness of the descriptor to these differences.
The reason RadVLAD performs well on the snowing sequence is because the sequence it trained on is the Bellmouth sequence from \textit{OORD} that similar noise state to the snowing sequence of Boreas.
Nevertheless, our method still maintains high performance and has the fastest processing time compared to other methods.

\begin{table}[h]
\caption{Overall Performance Evaluation in Diverse Weather}
\centering\resizebox{0.49\textwidth}{!}{
\begin{tabular}{c|l|cccc}
\toprule
\textbf{Sequence}                   & \textbf{Approach}          & \textbf{Recall@1 $\uparrow$}    & \textbf{F1 max $\uparrow$}       & \textbf{AUC $\uparrow$}      & \textbf{Proc. Time [s] $\downarrow$} \\ \hline
\multirow{5}{*}{Sunny}              & Radar SC                   & 0.815                & 0.961                 & \textbf{0.994}    &  0.389 \\
                                    & RaPlace                    & 0.816                & 0.905                 & 0.973             &  0.454 \\
                                    & RadVLAD                 & 0.778                & 0.899                 & 0.959             &  \underline{0.191} \\
                                    & \underline{FFT-RadVLAD}  & \underline{0.946}    & \underline{0.973}     & 0.981             &  0.198 \\
                                    & \textbf{ReFeree (Ours)}    & \textbf{0.967}       & \textbf{0.982}        & \underline{0.993}    &  \textbf{0.072} \\ \hline
\multirow{5}{*}{Snowing}            & Radar SC                   & 0.520                & 0.971                 & 0.554             &  0.380 \\
                                    & RaPlace                    & 0.848                & \underline{0.982}     & 0.996             &  0.507 \\
                                    & \textbf{RadVLAD}        & \underline{0.862}    & \textbf{0.992}        & \textbf{0.999}    &  \underline{0.222} \\
                                    & FFT-RadVLAD              & 0.480                & 0.964                 & 0.994             &  0.229 \\
                                    & \underline{ReFeree (Ours)} & \textbf{0.886}       & 0.969                 & \underline{0.997} &  \textbf{0.073} \\  \hline
                                    & \underline{Radar SC}       & 0.618                & \underline{0.921}     & \textbf{0.981}    &  0.384 \\
          Sunny                     & RaPlace                    & 0.013                & 0.176                 & 0.099             &  0.537 \\
            to                      & RadVLAD                 & \underline{0.806}    & 0.889                 & 0.959             &  \underline{0.264} \\
          Snowing                   & FFT-RadVLAD              & 0.033                & 0.081                 & 0.032             &  0.269 \\
                                    & \textbf{ReFeree (Ours)}    & \textbf{0.901}       & \textbf{0.948}        & \underline{0.961} &  \textbf{0.073} \\  \bottomrule

\end{tabular}}
\label{tab:boreas_overall}
\end{table}
\begin{figure}[h]
    \centering
    \def\width{0.47\textwidth}
    \includegraphics[clip, trim= 190 0 260 0, width=\width]{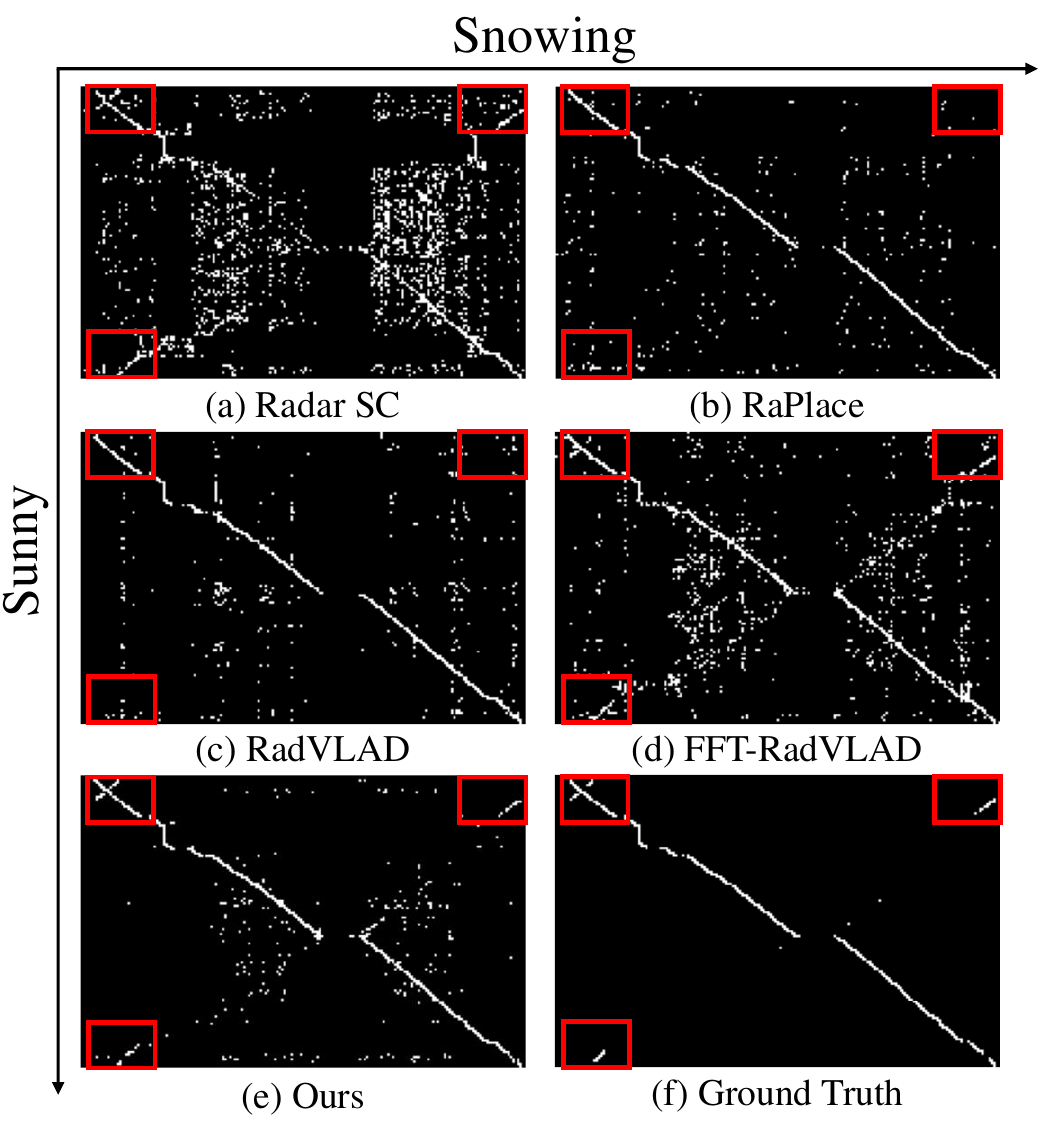}
    \vspace{-0.3cm}
    \caption{A embedding distance matrix is composed of the closest distance between a query (sunny sequence) and candidates from the map (snowing sequence).
             GT is the distance between the actual position of two poses, Radar SC is the cosine distance, RaPlace is the cross-correlation and auto-correlation, RadVLAD, FFT-RadVLAD, and our method utilizes L2 distance. 
             The values in the distance matrix for each descriptor represent the index of the closest embedding distance to their respective distance as 1 and the rest as 0.
    }
     \label{fig:heatmap}
     \vspace{-0.2cm}
\end{figure}
\subsubsection{Embedding Distance Matrix} 
As shown in \figref{fig:heatmap}, we constructed an embedding distance matrix by utilizing the closest distance with similar distances between the query (sunny sequence) and candidates from the map (snowing sequence).
The artifacts seen in \figref{fig:heatmap}(a), (b), (c), (d), (e) are caused by the incorrect loop detection that the query is far from the candidate on metric distance.
This phenomenon, which \citet{gadd2024oord} calls “perceptual ambiguity” or “aliasing”, occurs at the least in our results compared to GT.
Also, the forward loop detection results along to diagonal direction of the distance matrix are most similar to GT.
The most notable point of the distance matrix is the reverse loop result, highlighted in the red square, which is perpendicular to the diagonal direction.
These reverse loop results in the red square are most clearly depicted in our method and most closely resemble the GT.

\subsection{SLAM}
\subsubsection{Pose Graph Optimization}
\begin{figure}[t]
	\centering
	\def\width{0.45\textwidth}%
    {
        \includegraphics[clip, trim= 100 20 100 50, width=\width]{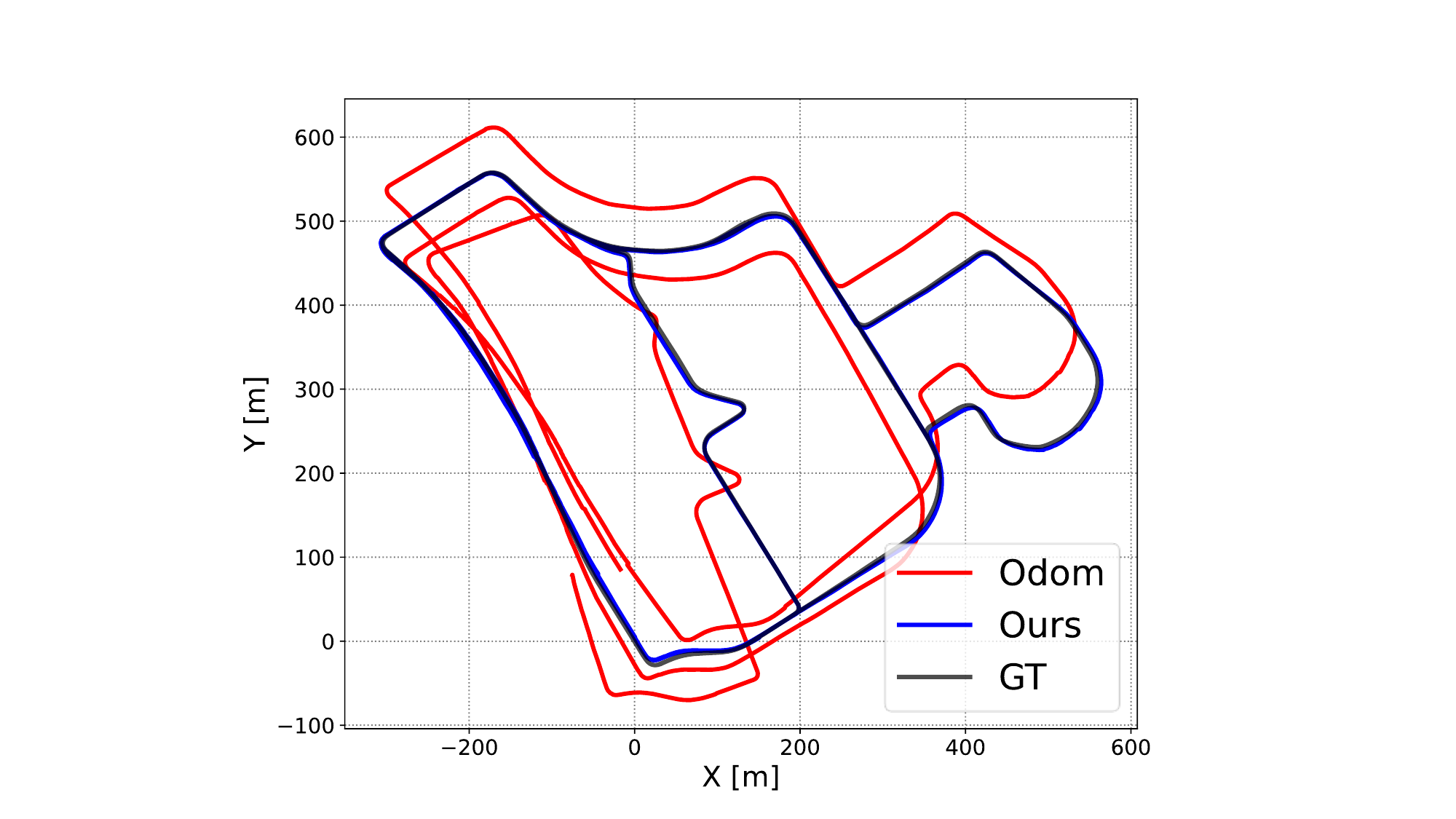}
    }
        \caption{A qualitative result of the trajectory with odometry, optimized pose, and ground truth in \textit{Mulran}'s \textit{KAIST} sequence. }
\label{fig:traj}
\end{figure}

As shown in \figref{fig:traj}, we compared the baseline odometry algorithm \cite{burnett_ral21} and the optimized pose through our SLAM pipeline with the ground truth pose from the EVO \cite{grupp2017evo} library.
Using our proposed methodology, we successfully demonstrate loop detection and closing in the pipeline.
As a result, the odometry, which diverged in the baseline algorithm, converged successfully to a trajectory similar to the ground truth.
\tabref{tab:evo} shows a quantitative evaluation APE according to the index.
Our method reduced APE about 50$\times$ compared to odometry.
\begin{table}[h]
\caption{Pose Error Evaluation in KAIST 03}
\centering\resizebox{0.49\textwidth}{!}{
\begin{tabular}{c|cccccc}
\toprule
              & \textbf{Max}    & \textbf{Mean}   & \textbf{Median} & \textbf{Min}   & \textbf{RMSE}   & \textbf{STD}    \\ \hline
Odometry      & 95.059          & 53.336          & 50.078          & 17.575         & 56.402          & 18.343 \\          \hline
\textbf{Ours} & \textbf{9.005}  & \textbf{3.493}  & \textbf{3.424}  & \textbf{0.048} & \textbf{3.912}  & \textbf{1.762}  \\ \bottomrule
\end{tabular}}
\vspace{-0.3cm}
\label{tab:evo}
\end{table}
\subsubsection{Processing Time}
\begin{figure}[h]
	\centering
	\def\width{0.49\textwidth}%
    {%
        \includegraphics[clip, trim= 0 95 0 85, width=\width]{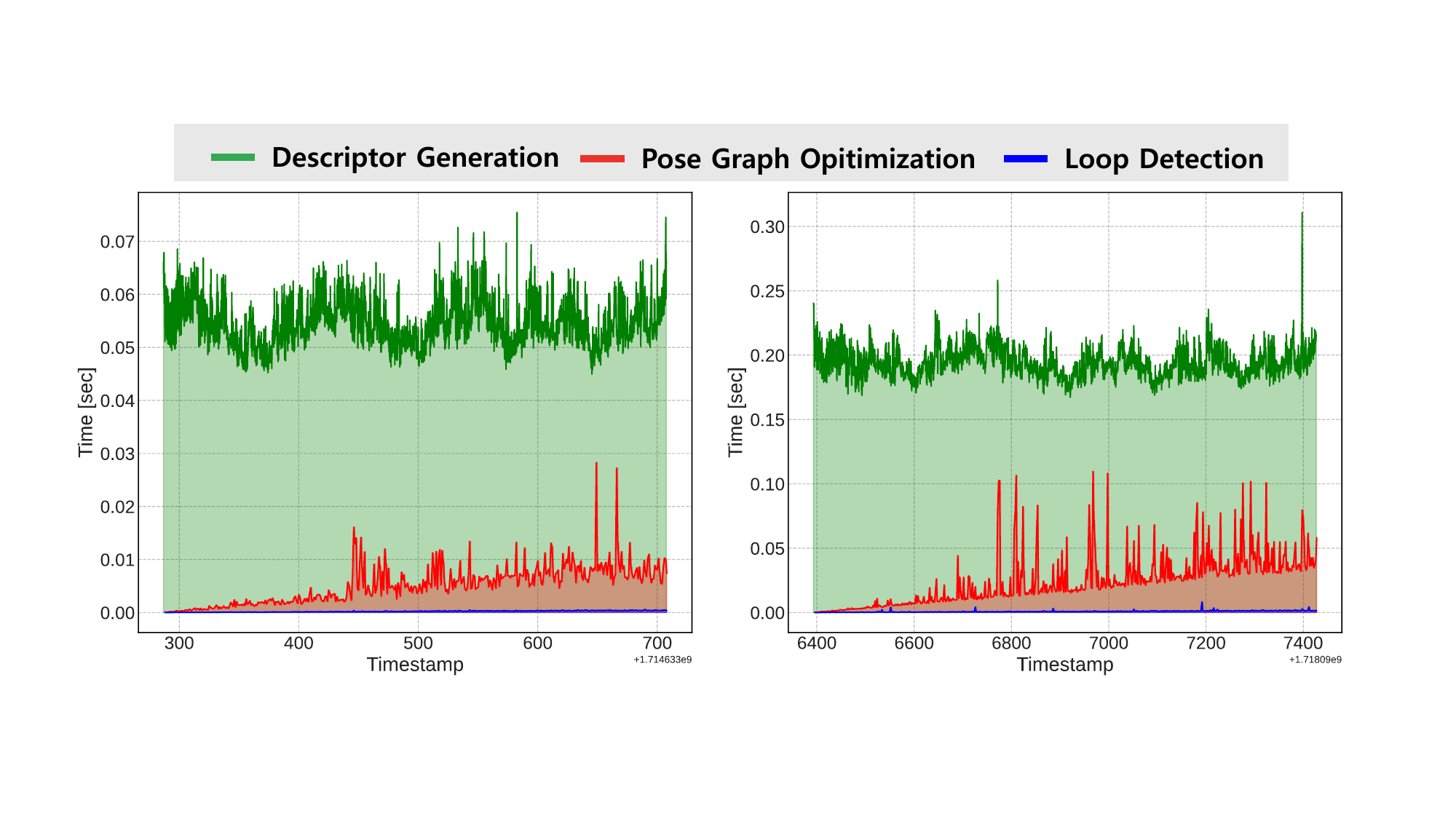}
    }
    \vspace{-0.6cm}
    \caption{Processing time on desktop (left) and Nvidia Jetson Nano (right). } 
\label{fig:time}
\vspace{-0.3cm}
\end{figure}

We measured the time of the SLAM pipeline on both desktop and Jetson Nano in \textit{DCC} of \textit{Mulran} sequence.
The results on the desktop are as shown on the left of \figref{fig:time}. Despite parallel processing, the descriptor generation process takes the most time due to the large size of the radar images.
\figref{fig:time}'s right shows the processing time on the Jetson Nano, indicating that this setup achieves a speed of approximately 4-5 Hz in an onboard environment.
Since radar sensor data operates at 4Hz, it is quite enough to certify the lightness of our descriptors and real-time performance.

\vspace{-0.2cm}

\subsection{Ablation Study}
We conducted an ablation study by applying radar-based feature algorithms to the free space. 
As represented in \tabref{tab:ablation}, these algorithms perform better than existing radar-based approaches. 
Also, the descriptors via free space all have similar performance, so it makes the most sense to use our methodology to improve the online performance on the onboard computer.
\begin{table}[h]
\caption{Overall performance of ReFeree with Different Features}
\centering\resizebox{0.49\textwidth}{!}{
{
\begin{tabular}{c|l|cccc}
\toprule
\textbf{Sequence}           & \textbf{Approach}                        & \textbf{Recall@1 $\uparrow$}    & \textbf{F1 max $\uparrow$}       & \textbf{AUC $\uparrow$}      & \textbf{Proc. Time [Hz] $\uparrow$}  \\ \hline
\multirow{3}{*}{DCC 01}     & Cen2018 \cite{cen2018precise}-ReFeree    & \textbf{0.723}         & 0.861                     & \underline{0.936}  &   \underline{11.86}\\
                            & Cen2019 \cite{cen2019radar}-ReFeree      & 0.684                  & \textbf{0.867}            & \textbf{0.938}     &  9.24\\ 
                            & Ours                                     & \underline{0.720}      & \textbf{0.867}            & \underline{0.936}  &  \textbf{15.15}\\ \hline
\multirow{3}{*}{KAIST 03}   & Cen2018 \cite{cen2018precise}-ReFeree    & \underline{0.953}      & \textbf{0.970}            & \textbf{0.991}     &  \underline{10.22}\\
                            & Cen2019 \cite{cen2019radar}-ReFeree      & 0.946                  & 0.946                     & 0.986              &  7.72 \\
                            & Ours                                     & \textbf{0.955}         & \textbf{0.970}            & \textbf{0.991}     &  \textbf{14.92}\\
\bottomrule

\end{tabular}}}
\label{tab:ablation}
\vspace{-0.3cm}
\end{table}

\section{Conclusion}
We proposed a radar-based lightweight and robust global descriptor using a feature and free space for place recognition.
We proved the proposed method's superiority through various experiments (i.e. single, multi, and different weather sessions). 
Additionally, by utilizing the properties of free space, we achieved place recognition even in extreme environments, and performed SLAM on the Jetson Nano with odometry and fast registration algorithms that suit our descriptor.
In future works, we will propose a radar-based distributed SLAM pipeline that highlights the advantage of lightness and additionally can be looked forward to having cross-modality with metrically free space measurable sensor.

%


\scriptsize
\bibliographystyle{packages/IEEEtranN} 
\bibliography{packages/string-short, packages/references}

\end{document}